\documentclass[11pt]{article}
\usepackage[margin=1in]{geometry}
\usepackage{amsmath,amssymb,amsfonts}
\usepackage{bm}
\usepackage{graphicx}
\usepackage{booktabs}
\usepackage{microtype}
\usepackage[numbers,sort&compress]{natbib}
\usepackage[colorlinks=true,linkcolor=blue,citecolor=blue,urlcolor=blue]{hyperref}

\newcommand{\tev}{t_{\mathrm{event}}}
\newcommand{\tpv}{t_{\mathrm{pv}}}
\newcommand{\ordOlmoOneCap}{0.71}
\newcommand{\ordOlmoOneCapT}{12}
\newcommand{\ordOlmoOnePre}{0.43}
\newcommand{\ordOlmoOnePreT}{8}
\newcommand{\ordOlmoTwoCap}{-0.14}
\newcommand{\ordOlmoTwoCapT}{21}
\newcommand{\ordOlmoTwoPre}{0.22}
\newcommand{\ordOlmoTwoPreT}{1}
\newcommand{\vCorpus}{466}

\newcommand{\vProspBar}{5{,}200}
\newcommand{\vProspLead}{18{,}000}
\newcommand{\vRfivebLead}{34{,}200}
\newcommand{\vRoneLead}{10{,}400}
\newcommand{\vRoneRel}{60\%}
\newcommand{\vShiftCos}{4{,}400}
\newcommand{\vShiftDcos}{16{,}400}
\newcommand{\vTwoGateLead}{12{,}400}
\newcommand{\vTwoGateRel}{53\%}
\newcommand{\wBareFA}{0/10}
\newcommand{\wConfCover}{15/15}
\newcommand{\wConfDelta}{975}
\newcommand{\wConfQ}{150}
\newcommand{\wConfWidth}{300}
\newcommand{\wConjFA}{0/10}
\newcommand{\wConjPre}{30/30}

\newcommand{\wEventMax}{7{,}400}
\newcommand{\wEventMin}{5{,}650}
\newcommand{\wFleetNeg}{15}
\newcommand{\wFleetPos}{30}

\newcommand{\wGapBatchRatio}{2.35}

\newcommand{\wGapLrRatio}{2.27}
\newcommand{\wGateCcover}{9/10}
\newcommand{\wGateCleadMed}{987.5}
\newcommand{\wGateCover}{10/10}
\newcommand{\wGateCpre}{10/10}
\newcommand{\wGateCrho}{0.952}
\newcommand{\wGateEvents}{10/10}
\newcommand{\wGateLeadMed}{1{,}012}
\newcommand{\wGatePre}{10/10}
\newcommand{\wGateRho}{0.988}
\newcommand{\wIndRho}{-0.03}
\newcommand{\wLawFrac}{0.843}
\newcommand{\wLawIQR}{0.825--0.865}
\newcommand{\wLawMult}{1.19}
\newcommand{\wLawN}{80}
\newcommand{\wLawRange}{0.739--0.898}
\newcommand{\wLeadMax}{1{,}125}
\newcommand{\wLeadMed}{975}
\newcommand{\wLeadMin}{825}
\newcommand{\wLossConfLead}{50}
\newcommand{\wLossLeadMed}{50}
\newcommand{\wLossLeadMin}{-25}
\newcommand{\wLossRho}{0.977}
\newcommand{\wLossTheta}{2.077}
\newcommand{\wNegConjFA}{0}
\newcommand{\wNegTotal}{33}
\newcommand{\wNineCover}{5/5}
\newcommand{\wNineEvents}{5/5}
\newcommand{\wNineMultRange}{1.160--1.226}
\newcommand{\wNorepPvRange}{0.036--0.061}
\newcommand{\wOnePrefixMax}{0.014}
\newcommand{\wPyCopyA}{0.022}
\newcommand{\wPyCopyB}{9.7}
\newcommand{\wPyPrefixA}{0.019}
\newcommand{\wPyPrefixB}{0.93}
\newcommand{\wPyPrevA}{0.37}
\newcommand{\wPySeventyLatePrefix}{0.36}
\newcommand{\wPythiaPreLeads}{1--8}
\newcommand{\wPythiaRuns}{5}
\newcommand{\wRho}{0.977}
\newcommand{\wRhoCIhi}{0.995}
\newcommand{\wRhoCIlo}{0.911}
\newcommand{\wSpread}{1.31}
\newcommand{\wTotalFleet}{45}
\newcommand{\wTrapBareFA}{10/10}
\newcommand{\wTrapConjFA}{0/10}
\newcommand{\wTrapConjLeadMed}{1{,}900}
\newcommand{\wTrapConjLeadRange}{1{,}525--2{,}100}
\newcommand{\wTrapConjRho}{1.000}
\newcommand{\wTrapIndFA}{2/10}
\newcommand{\wTrapNegPvRange}{650--4{,}400}
\newcommand{\wTrapPrefixLead}{550}
\newcommand{\wTrapPrefixRho}{0.988}

\title{Capability Emergence Can Be Forecast:\\
\large Per-Seed, In Advance, With Calibrated Intervals, Certified False Alarms,\\
and a Blind Pre-Registered Gate}
\author{Gunner Levi Howe\\ \small{\texttt{gunnerlevihowe@gmail.com}}}
\date{July 2026}

\begin{document}
\maketitle

\begin{abstract}
Emergent capabilities are widely treated as unpredictable: aggregate loss improves
smoothly while specific abilities appear abruptly. Prior work offers early-warning
\emph{indicators} but never scores them as \emph{forecasts} --- no lead time at a
controlled false-alarm rate, no calibration, no negative controls, no blind validation.
We supply that discipline and show that, in controlled settings spanning grokking model
systems and small language models, emergence \emph{timing} is forecastable per training
run, in advance, with calibrated uncertainty. On language models: across \wFleetPos{}
transformers at identical configuration --- where the strongest published baseline
predicts a single constant \citep{aoyama2025predicting} --- the formation time of the
previous-token head (the mechanistic precursor of induction
\citep{olsson2022context}) forecasts each seed's induction emergence at Spearman
$\rho=\wRho$ (95\% CI [\wRhoCIlo, \wRhoCIhi]) with median lead \wLeadMed{} steps
(${\sim}15\%$ of training). A best-case training-loss rule \emph{ties} that ranking
($\rho=\wLossRho$) but with \wLossLeadMed-step median lead --- a nowcast, not a forecast.
Split-conformal intervals of width \wConfWidth{} steps, anchored at the precursor alarm,
covered \wConfCover{} held-out seeds at 90\% nominal. The frozen rule then passed blind
pre-registered gates on \emph{two} never-seen configurations --- a wider model with
sparser repetition (\wGateCover{} interval coverage, bar 7/10) and an entirely different
language (\wGateCcover, exactly nominal) --- with zero false alarms throughout. A
trap-language rung then attacked our own rule as pre-registered: in a language where
previous-token context pays for the task itself, the bare precursor false-alarms on
\wTrapBareFA{} capability-blocked runs, while the mechanism-composed conjunction is
certified in \emph{both} language classes (0 false alarms) and times emergence at
$\rho = \wTrapConjRho$. Finally, a gap-origin study broke the fixed offset itself (both
learning rate and batch size move the gap ${\sim}2.3\times$; no external clock owns it)
and revealed the scale-invariant law beneath: across \wLawN{} valid-anchor runs spanning
every fleet, both languages, and all configuration axes, the anchor fires at
\wLawFrac{} of time-to-emergence (IQR \wLawIQR) --- $\tev \approx \wLawMult \times
t_{\mathrm{anchor}}$ --- and this multiplicative rule passed its own blind gate at a
third never-seen configuration (\wNineCover{} coverage). All false-alarm claims are
certified against \wNegTotal{} \emph{manufactured capability-blocked negatives}
(data-ablated and architectural, across both languages) --- a resource we argue is mandatory for capability
monitoring and that no public suite provides. Public-checkpoint evidence: the precursor
leads the capability across \emph{three model families} (Pythia, OLMo, OLMo-2; seven
suites, pre-registered rules) --- OLMo-2's 1B-token checkpoint shows the precursor
formed while the capability is absent --- yet all Pythia suites cliff at the same token
count, confirming that public artifacts cannot test per-seed forecasting. The grokking half of the paper (\vCorpus{} runs) contributes the evaluation
methodology and its hard lessons: a robustness-versus-false-alarm tradeoff,
mechanism-factored gates that resolve it, non-transfer of fitted calibration, and a
label-free probe family that died under pre-registered confirmation. Four pre-registered
kill criteria fired across the program and are reported. Every prediction, freeze, and
verdict is commit-stamped before its data existed. Scope: named capabilities with known
mechanistic precursors, small models; we state what stands between this and frontier
deployment.
\end{abstract}

\section{Introduction}

When a capability emerges during training, the training curve rarely announces it in
advance: loss declines smoothly while the ability arrives abruptly
\citep{olsson2022context,schaeffer2023mirage}. For frontier-scale systems this
unpredictability is a safety problem with a name --- labs want to know what a model will
be able to do \emph{before} it can do it. A literature of early-warning signals has
grown in response: loss-curve oscillations that predict whether grokking will occur
\citep{notsawo2023predicting}, spectral observables that show precursors of phase
transitions \citep{hennick2026density}, geometric stage detection
\citep{hoogland2024developmental}, and scale-axis extrapolations
\citep{hu2024passuntil,snell2024predicting}. What none of this work does is score a
\emph{forecast}: state, in advance, when a capability will arrive; attach calibrated
uncertainty; measure lead time at a controlled false-alarm rate against negatives where
the capability never arrives; and validate the frozen system blind. Indicators without
that discipline are weather lore, not weather forecasting.

\begin{figure}[t]
\centering
\includegraphics[width=0.72\textwidth]{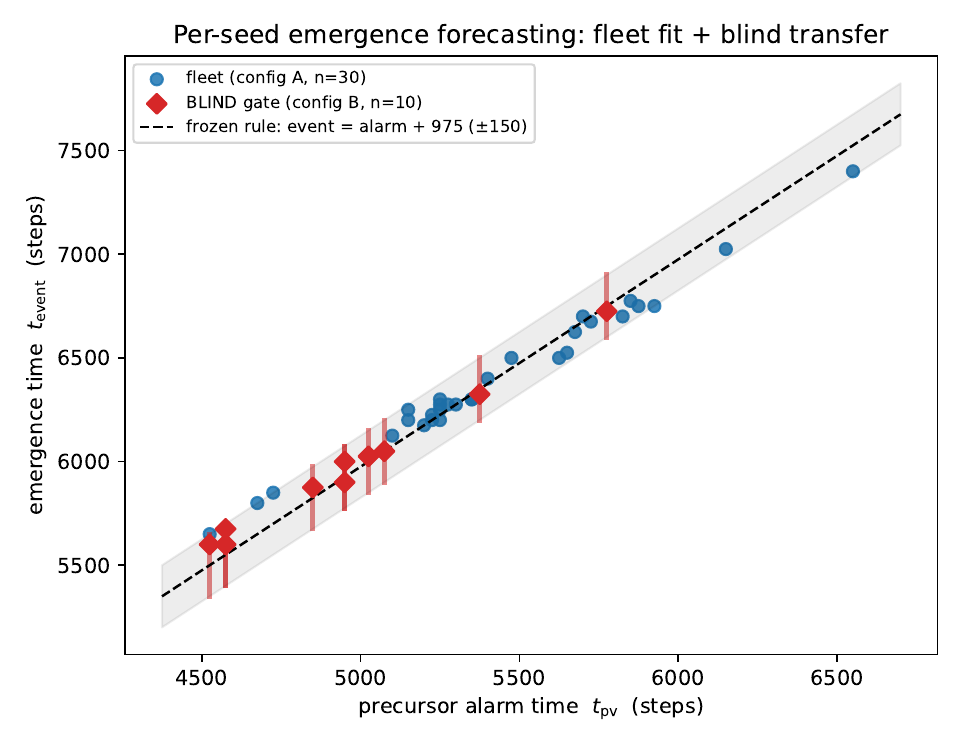}
\caption{\textbf{The central result.} Precursor alarm time vs.\ emergence time. Blue:
the calibration fleet (30 seeds, config A) --- Spearman $\rho=\wRho$. The dashed line
and band are the \emph{frozen} rule (event $=$ alarm $+\,\wConfDelta \pm \wConfQ$).
Red diamonds with bars: the blind gate --- 10 fresh seeds on a never-seen configuration
(wider model, sparser repetition), scored once against the frozen rule: \wGateCover{}
covered. Negatives (18 manufactured capability-blocked runs, not shown) produced zero
false alarms.}
\label{fig:forecast}
\end{figure}

This paper does two things. \textbf{First}, it builds the missing evaluation discipline
on a grokking model system --- \vCorpus{} training runs across two domains with
interventions that shift emergence timing ${\sim}100\times$ --- and reports what the
discipline kills: single-signal alarms die on look-alike negatives; the context features
that control false alarms are exactly the features that break under interventional
shift (a robustness--false-alarm tradeoff); fitted calibration does not transfer across
regimes; and an initially spectacular label-free probe family failed its pre-registered
confirmation. \textbf{Second}, it carries the surviving methodology to language models
and obtains the result the discipline was built for
(Figure~\ref{fig:forecast}): per-seed forecasts of induction-head emergence from a
mechanistic precursor, with calibrated intervals, certified false alarms, and a blind
pre-registered transfer gate passed at ceiling.

Contributions, each with its adversarial control:
\begin{enumerate}
\item \textbf{A forecasting benchmark and protocol for emergence timing}: event
  definitions robust to outliers, lead time at capped false-alarm rate, champion
  selection on training data only, (correlation, lead) reported as a pair --- plus a
  \emph{negative taxonomy} (budget-censored vs.\ structurally blocked vs.\
  \emph{manufactured}) and a minimum-negative-count rule learned the hard way.
\item \textbf{The per-seed result} (Sections~\ref{sec:fleet}--\ref{sec:conformal}): at
  fixed configuration, where the config-equation baseline of
  \citet{aoyama2025predicting} is a constant by construction, precursor formation time
  forecasts emergence at $\rho=\wRho$ with \wLeadMed-step median lead; a best-case loss
  rule ties the correlation with \wLossLeadMed{} steps of lead. Warning time lives in
  \emph{where you anchor}, and only the circuit precursor anchors early.
\item \textbf{Manufactured negatives, and the trap that proves they matter}
  (Sections~\ref{sec:fleet}, \ref{sec:trap}): capability-blocked training runs ---
  data-ablated and architectural --- enabling the first certified false-alarm rates for
  emergence forecasting. In simple languages the bare precursor is certified
  (\wBareFA{} FA); in a trap language built so previous-token context pays for the task
  itself, it fails completely (\wTrapBareFA{} FA on blocked runs, as pre-registered) and
  the mechanism-composed conjunction is certified in both regimes (0 FA) while timing
  emergence at $\rho = \wTrapConjRho$. Anchors must be composed from mechanism, not read
  off a single circuit.
\item \textbf{Three blind gates and a scale-invariant law} (Sections~\ref{sec:gate},
  \ref{sec:law}): the frozen rule passed pre-registered blind gates on two never-seen
  configurations (\wGateCover{} and \wGateCcover{} coverage); a gap-origin study then
  broke the fixed offset (both optimization axes move the gap; no external clock owns
  it) and exposed the law beneath --- the anchor fires at \wLawFrac{} of
  time-to-emergence across \wLawN{} runs --- whose multiplicative form passed a third
  blind gate (\wNineCover) at yet another unseen configuration.
\item \textbf{The kill ledger} (Section~\ref{sec:kills}): four pre-registered kill
  criteria fired across the program and are reported with their mechanisms, alongside
  every failed prediction. The positive results above survived the same regime that
  produced these deaths; that, not any single number, is the paper's claim to trust.
\end{enumerate}

Everything is pre-registered in a public commit chain (every freeze precedes its data;
Section~\ref{sec:repro}), and the corpus, fleets, and harness are released as a
benchmark.

\section{Related work}
\label{sec:related}

\citet{notsawo2023predicting} predict \emph{whether} grokking will occur from early
loss-curve oscillations --- a scalar whether-classifier; its oscillation feature is one
of our baselines. \citet{hennick2026density} derive spectral early-warning indicators
(2-RDM heat capacity) across four transition settings but report no lead time,
calibration, false-alarm rate, or baseline race; our label-free rung
(Section~\ref{sec:kills}) shows concretely why unscored spectral indicators can
mislead: ours died under confirmation. \citet{hoogland2024developmental} detect
developmental stages retrospectively via the local learning coefficient.
\citet{aoyama2025predicting} predict induction-head formation time from pre-training
config (batch, context, bigram statistics) --- powerful, but constant per config; our
fleet result answers precisely the variance their equation cannot (per-seed timing at
fixed config), and our Pythia analysis confirms their account across public suites (all
five cliff at the same token count). Scale-axis emergence prediction
\citep{hu2024passuntil,snell2024predicting} forecasts across model size, orthogonal to
our within-run time axis. \citet{olsson2022context} established the induction-head
phase change and its previous-token precursor; \citet{barak2022hidden} and progress
measures \citep{nanda2023progress} established hidden progress; we turn these
observations into scored forecasts. The grokking substrate builds on
\citep{power2022grokking,liu2023omnigrok,weightnormdelay2026} and our companion papers
\citep{companion2026a,companion2026b,companion2026c}, which established the
representational-timing mechanism and the norm clock that our two-gate forecaster
exploits.

\section{Forecasting emergence: definitions and discipline}
\label{sec:defs}

A \textbf{run} logs behavioral and internal signals at fixed cadence. An \textbf{event}
is the first crossing of an absolute behavioral criterion robust to outliers (never a
fraction of a run's own maximum --- a rule adopted after that operationalization failed
twice; Section~\ref{sec:kills}). A \textbf{forecaster} observes the log prefix and may
\textbf{alarm} once; its \textbf{lead} is $\tev - t_{\mathrm{alarm}}$ (alarms after the
event score zero; no alarm on an eventing run is a miss). \textbf{False-alarm (FA)
rate} is the fraction of \emph{negative} runs --- runs where the capability never
arrives --- that alarm. Thresholds and champions are selected on training runs only,
under an FA cap ($\leq 5\%$), with at least five negatives on every side of every
split (a minimum learned from a degenerate fit, Section~\ref{sec:kills}). Negatives
come in three kinds: \emph{budget-censored} (would event with more budget --- these
legitimately attract alarms), \emph{structurally blocked} (wrong learned structure),
and \emph{manufactured} (constructed so the capability cannot form). Forecast quality
is always a \emph{pair} --- (ranking skill, lead) --- because a rule can rank
perfectly by detecting the event as it happens; and point forecasts carry
split-conformal intervals whose empirical coverage is reported. Blind validation means:
every constant frozen and commit-stamped, then fresh runs, one scoring pass.

\section{Part I: the grokking benchmark, and what the discipline kills}
\label{sec:grok}

We built the protocol on \vCorpus{} pre-existing grokking runs (modular arithmetic
transformers; MNIST MLPs in the Omnigrok regime) whose companion-paper interventions
(contrastive priors, norm clamps) shift emergence timing up to ${\sim}100\times$ and
--- crucially --- \emph{decouple} the dominant scalar clock (weight norm) from the
outcome. Full protocol and verdicts are in the repository
(\texttt{plan\_p5.md}, \texttt{P5\_RESULTS.md}); four results shape everything
downstream.

\begin{figure}[t]
\centering
\includegraphics[width=0.62\textwidth]{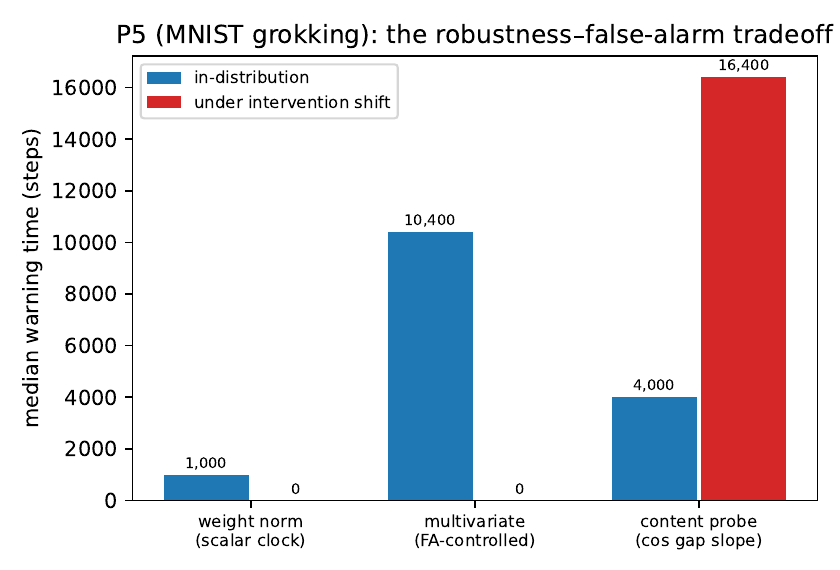}
\caption{\textbf{The robustness--false-alarm tradeoff} (MNIST grokking; median warning
time at $\leq 5\%$ FA, higher is better). The scalar clock (weight norm) forecasts
in-distribution but collapses when interventions decouple the norm from the outcome.
The multivariate forecaster that best controls false alarms in-distribution
(\vRoneLead{} steps, \vRoneRel{} of the delay) collapses hardest under shift --- its
norm-context features are exactly what the intervention breaks. The bare content probe
survives the shift (\vShiftDcos{} steps) but is FA-fragile in corpora with look-alike
negatives.}
\label{fig:tradeoff}
\end{figure}

\textbf{(i) In-regime forecasting works.} A multivariate forecaster with
mechanism-motivated features achieved \vRoneLead-step median lead (\vRoneRel{} of the
delay) on held-out seeds at zero structural false alarms, and \vProspLead{} steps
prospectively on pre-registered fresh runs including a never-seen norm pin (bar
\vProspBar) --- all four prospective predictions passed.

\textbf{(ii) The robustness--false-alarm tradeoff}
(Figure~\ref{fig:tradeoff}). Trained on control runs only and tested on
intervention arms, the scalar norm clock collapses (median lead 0; the interventions
hold the norm where controls never generalize), and the FA-controlled multivariate
collapses with it --- \emph{because} the context features that suppress false alarms
are the features the intervention breaks. The bare content probe survives
(\vShiftCos--\vShiftDcos{} steps). A monitor tuned quiet on its training distribution
goes blind exactly when the recipe changes.

\textbf{(iii) Mechanism-factored gates resolve the tradeoff where mechanism is known.}
Composing a content gate with an \emph{a-priori} norm-viability window (constants from
the companion papers, not fit) retains \vTwoGateLead{} steps (\vTwoGateRel{} of the
delay) under the same shift at zero miss. The portable ingredient is mechanism, not
fitted calibration: fitted thresholds transferred across domains in neither direction
(kill K5).

\textbf{(iv) The discipline kills seductive numbers.} An algorithmic-domain corpus
with structure-complete-but-blocked negatives admitted \emph{no} forecaster at 5\% FA
(kill K1/K8): trap negatives carry more ``progress'' signal than true positives show
before their events. And a label-free spectral probe that posted a \vRfivebLead-step
validated lead died under pre-registered confirmation --- it false-alarmed on fresh
structural negatives and, per-run, was largely detecting the \emph{intervention}, not
the emergence (kill K9). Both deaths are in the ledger (Section~\ref{sec:kills}) and
both shaped the LM protocol.

\section{Part II: language models}

\subsection{Public checkpoints: the precursor leads, and why public suites are not
enough}
\label{sec:pythia}

\begin{figure}[t]
\centering
\includegraphics[width=0.98\textwidth]{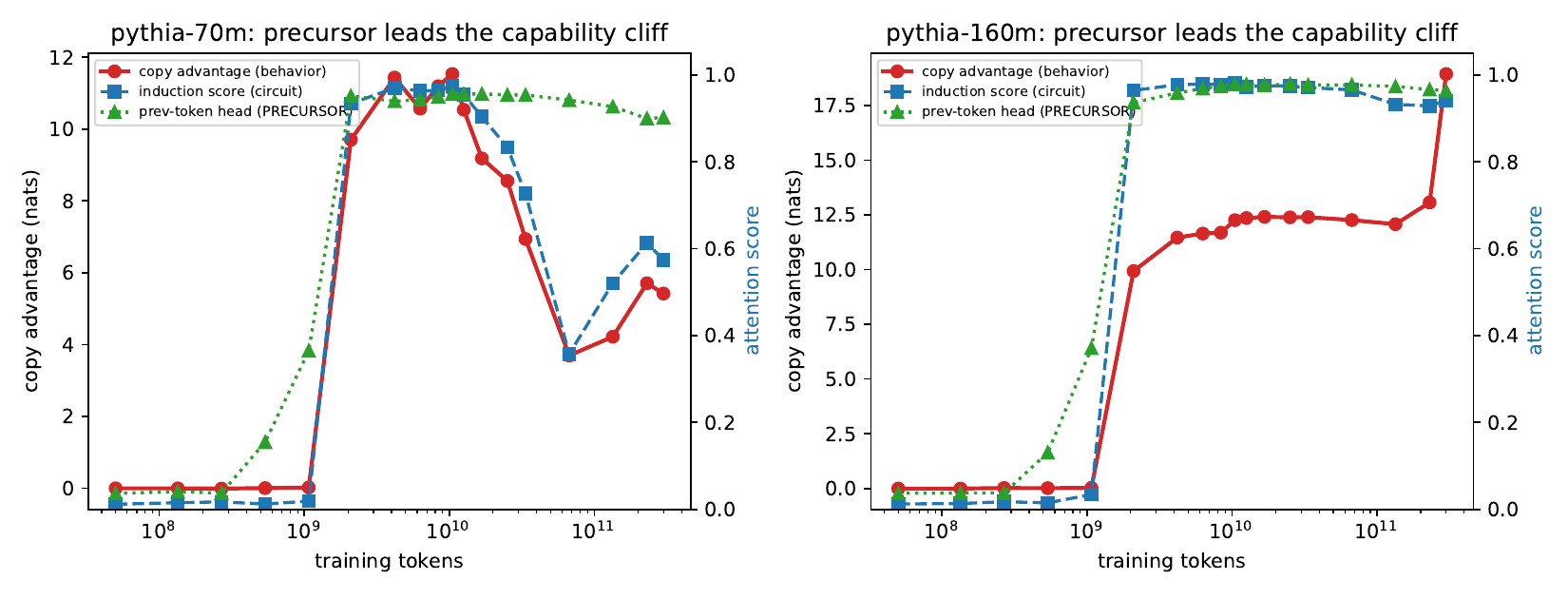}
\caption{\textbf{Public Pythia checkpoints (models we did not train).} The
previous-token head (green) rises a full public-checkpoint stage before the
induction-head score (blue) and the behavioral copy advantage (red) cliff together.
Left: pythia-70m; right: pythia-160m. The 70m model also partially \emph{loses}
induction late in training (prefix \wPyPrefixB$\to$\wPySeventyLatePrefix{} at 67B
tokens) --- capability regression exists and is monitorable.}
\label{fig:pythia}
\end{figure}

We probed \wPythiaRuns{} Pythia suites \citep{biderman2023pythia} (70m/160m/410m,
standard and deduped) across training checkpoints for three quantities: behavioral copy
advantage on repeated random sequences, the maximum per-head prefix-matching (induction)
score, and the maximum early-layer previous-token score --- the known mechanistic
precursor \citep{olsson2022context}. The phase change is crisp in every suite
(Figure~\ref{fig:pythia}): at 1.07B tokens copy advantage is \wPyCopyA{} and the
precursor is already at \wPyPrevA{} (${\sim}10\times$ baseline) while induction sits at
\wPyPrefixA; one stage later both cliff (copy \wPyCopyB{} nats, induction
\wPyPrefixB). Against a best-case loss threshold granted an oracle sweep, the precursor
alarmed strictly earlier on 4/5 suites and never post-event (leads \wPythiaPreLeads{}
stages); our pre-registered prediction here failed in both directions --- the precursor
led by \emph{more} than predicted and loss was \emph{stronger} than predicted (one
stage of genuine timing) --- and is reported as registered. Two structural facts matter
more than the scores. First, all five suites, which share batch and context
configuration, cliff at the same ${\sim}2.1$B tokens --- consistent with
config-determined timing \citep{aoyama2025predicting} and fatal to using public suites
for per-seed forecasting: there is one run per config. Second, an event definition we
had frozen (50\% of each run's maximum) was silently broken by a late-training outlier
in one suite --- the same operationalization failure our grokking rungs had already
punished --- and was replaced, in print, by absolute criteria everywhere.

\paragraph{Cross-family replication (pre-registered).} To test whether the
precursor-leads phenomenon is a Pythia artifact, we froze the same rules (precursor
$\geq 0.10$; capability $\geq 2.0$ nats) and swept two families this program had never
probed --- AllenAI's two OLMo generations, with architectures, data pipelines, and
tokenizers all distinct from Pythia's and from each other. All three pre-registered
predictions passed in \emph{both} families, and each family's public grid caught the
intermediate state directly: OLMo-1 at 8B tokens shows the precursor formed
(\ordOlmoOnePre) with the capability below threshold (\ordOlmoOneCap{} nats), and OLMo-2
at 1B tokens shows the precursor formed (\ordOlmoTwoPre) with the capability
\emph{absent} (\ordOlmoTwoCap{} nats) --- ordinal leads of
\ordOlmoOnePreT{}$\to$\ordOlmoOneCapT{}B and \ordOlmoTwoPreT{}$\to$\ordOlmoTwoCapT{}B
tokens respectively. The kill (capability before precursor in any family) fired
nowhere. Across three model families and seven public suites, the precursor leads.

\subsection{The seed fleet: per-seed forecasts, certified against manufactured
negatives}
\label{sec:fleet}

Public suites cannot answer the forecasting question; fleets can. We trained
\wTotalFleet{} two-layer transformers on a fixed synthetic language (bigram-structured,
with variable-offset within-context repetition so that positional shortcuts cannot
substitute for induction --- a shortcut our first design permitted, caught in smoke
testing): \wFleetPos{} positives at one configuration (seeds vary; config fixed), plus
\wFleetNeg{} \textbf{manufactured negatives} --- 10 \emph{data-ablated} (identical
language, zero repetition: induction has no training signal) and 5 \emph{architectural}
(one layer: induction is structurally impossible \citep{olsson2022context}). Events
use an absolute criterion (copy advantage $\geq 2$ nats, two consecutive evals).

Results, scored once against the pre-registered spec. Per-seed event times span
\wEventMin--\wEventMax{} steps (spread \wSpread$\times$; the pre-registered bar of
$1.3\times$ was cleared by 0.01 --- variance at fixed config is real but modest, and we
do not dramatize it). The precursor's crossing time forecasts each seed's event:
$\rho = \wRho$ (bootstrap CI [\wRhoCIlo, \wRhoCIhi], $n=\wFleetPos$), median lead
\wLeadMed{} steps (\wLeadMin--\wLeadMax) --- ${\sim}15\%$ of the run issued as advance
notice, seed by seed, where the config equation predicts one constant. The best-case
loss threshold ($\theta = \wLossTheta$, granted an oracle sweep) \emph{ties} the
ranking ($\rho=\wLossRho$) --- and forecasts nothing: its median lead is
\wLossLeadMed{} steps (minimum \wLossLeadMin), i.e., it detects the cliff in progress.
Our pre-registered comparison metric (rank correlation alone) was therefore
\emph{wrong as designed}; the two-dimensional truth --- equal correlation, $20\times$
the lead --- is reported alongside the registered failure, and (correlation, lead)
pairs are now the protocol's required form. A second dissociation: the graded
\emph{behavioral} ramp that precedes the cliff carries no timing information
($\rho = \wIndRho$) --- circuits time the transition; early behavior does not.

The negatives certified the false-alarm side. On data-ablated runs the precursor never
formed (max layer-0 previous-token score \wNorepPvRange{} vs.\ threshold 0.10): bare
precursor FA \wBareFA, conjunction FA \wConjFA, with pre-event alarms on \wConjPre{}
positives. Here our pre-registered mechanistic prediction was wrong and is reported:
we expected the precursor to form anyway (predicting false alarms that would require a
conjunctive gate); in fact a pure bigram language is solvable through the embedding
pathway alone, so previous-token attention only pays where repetition exists. The
manufactured negatives thus \emph{certified} the precursor rather than trapping it ---
certification is the other thing negative sets are for --- and constructing a language
where the trap is real (where previous-token context pays independently of repetition)
is the named follow-up. Architectural negatives: zero events, induction score
$\leq \wOnePrefixMax$ throughout, zero alarms.

\subsection{Calibrated intervals}
\label{sec:conformal}

Single thresholds are fragile (a P5 lesson bought with a dead result); forecasts should
be intervals with measured coverage. Split-conformal calibration (seeds 1--15) around
the anchored point forecast $\tpv + \wConfDelta$ yields intervals of median width
\wConfWidth{} steps that covered \wConfCover{} held-out seeds (nominal 90\%) --- issued
\wConfDelta{} steps before the event. The identical machinery anchored at the loss
crossing covers too --- with \wLossConfLead{} steps of lead. Anchoring, not calibration,
is where warning time comes from.

\subsection{Two blind gates}
\label{sec:gate}

Everything above is one configuration. The ship-gate: freeze the entire rule verbatim
--- alarm at layer-0 previous-token score $\geq 0.10$; forecast the event in
$[t_{\mathrm{alarm}} + 825,\; t_{\mathrm{alarm}} + 1125]$ --- commit it, then score once
on fresh runs the rule never saw. \textbf{Gate B} (model width $256 \to 320$; repetition
density $0.75 \to 0.6$; 10 seeds + 3 negatives): \wGateEvents{} events, \wGatePre{}
pre-event alarms, \textbf{\wGateCover{} interval coverage} (bar 7/10), $\rho =
\wGateRho$, 0/3 false alarms, median lead \wGateLeadMed{} steps. \textbf{Gate C}
(the language itself replaced --- a different generator seed produces a different
bigram table; original architecture): \wGateCcover{} coverage --- \emph{exactly} the
90\% nominal the conformal quantile promises --- \wGateCpre{} pre-event, 0/3 false
alarms (bare rule also 0/3), $\rho = \wGateCrho$, median lead \wGateCleadMed. Our
pre-registered expectation was humble both times: the P5 transfer kill made the frozen
offset the component we flagged as most at risk. It survived both shifts --- and
Section~\ref{sec:law} explains why, less flatteringly than we had hoped.

\subsection{The trap language: our rule, attacked as pre-registered}
\label{sec:trap}

The fleet's negatives certified the bare precursor partly because a pure bigram language
gives previous-token attention no task role (Section~\ref{sec:fleet}). So we built the
language where it has one --- next-token distributions selected by the previous token
(one learnable bit of pair context) --- and pre-registered the attack on our own
headline rule: the precursor should now form \emph{everywhere}, including where induction
cannot. It does. On 10 capability-blocked runs in the trap language the bare precursor
rule false-alarms \textbf{\wTrapBareFA} (crossing threshold at \wTrapNegPvRange{} steps
from the task incentive alone), and on 2/10 positives the graded behavioral signal
false-alarms transiently on its own (\wTrapIndFA). Neither gate survives alone. The
\emph{same-eval conjunction} of the two --- frozen a priori in the pre-registration ---
is certified in both language classes (\wTrapConjFA{} FA here; \wConjFA{} in the bigram
fleet) and, remarkably, times emergence \emph{better} than the bare precursor ever did:
$\rho = \wTrapConjRho$ across 10 seeds with median lead \wTrapConjLeadMed{} steps
(range \wTrapConjLeadRange). A secondary anchor --- the induction score's own early ramp
--- gives $\rho = \wTrapPrefixRho$ at \wTrapPrefixLead-step lead, also 0 FA. Design
iterations disclosed: two earlier trap constructions failed instructively (an
unlearnable language gives the precursor no incentive at all; too rich a language starves
induction past budget), and both feed Section~\ref{sec:law}'s account.

\subsection{From offset to law: the gap is a fraction, not a clock}
\label{sec:law}

\begin{figure}[t]
\centering
\includegraphics[width=0.98\textwidth]{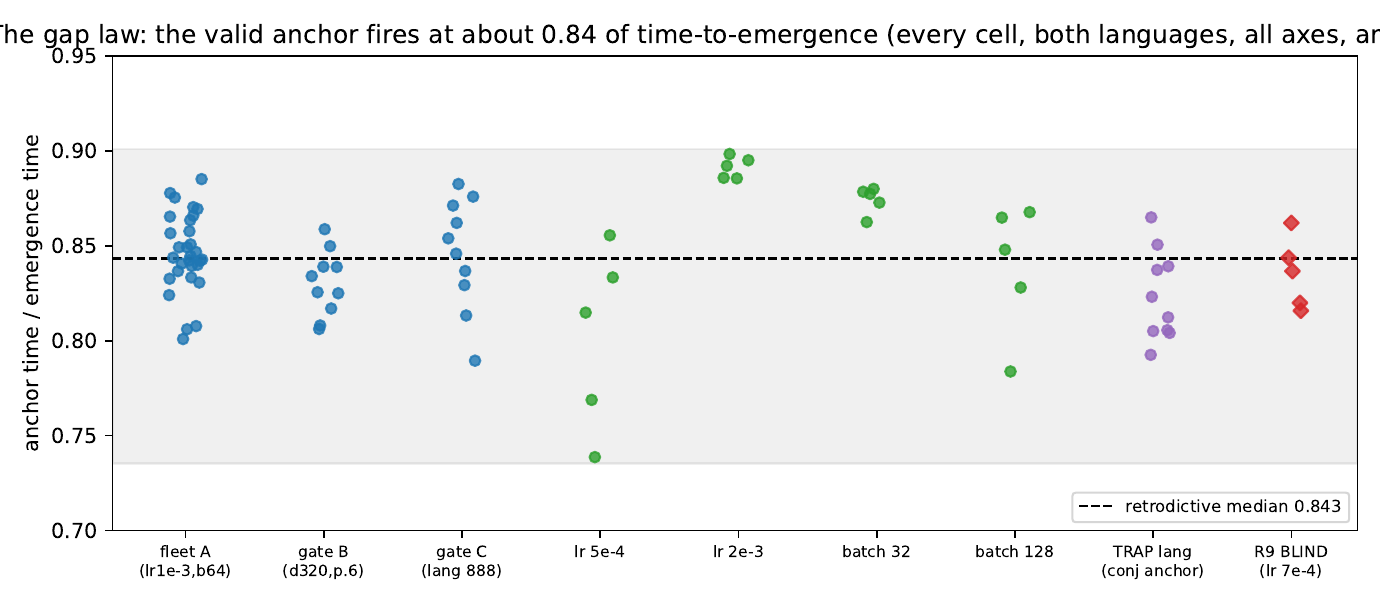}
\caption{\textbf{The gap law.} Anchor time over emergence time for every valid-anchor
cell in the program: the calibration fleet, both blind gates, all four gap-probe cells
(learning rate halved/doubled, batch halved/doubled), the trap language via its
conjunction anchor, and --- red diamonds --- the blind validation at a never-seen
learning rate. Median \wLawFrac{} (dashed); the shaded band is the frozen envelope
$[1/1.36,\,1/1.11]$ used in the blind test. One fraction describes \wLawN{} runs whose
emergence times span $3\times$.}
\label{fig:gaplaw}
\end{figure}

Why did a fixed \wConfDelta-step offset survive two config shifts? The pre-registered gap-origin
study answers by breaking it. Halving the learning rate stretches the gap
$\wGapLrRatio\times$; halving the batch stretches it $\wGapBatchRatio\times$: both
optimization axes move it, so \emph{no external clock owns the gap} (the registered
discrimination rule finds no owner; kill K8b fires as written). The post-hoc analysis
--- labeled as such --- finds the invariant underneath (Figure~\ref{fig:gaplaw}): across
all \wLawN{} valid-anchor runs in the program, spanning both languages, every
configuration axis, and a $3\times$ range of emergence times, the anchor fires at
\wLawFrac{} of time-to-emergence (IQR \wLawIQR, range \wLawRange). The gates of
Section~\ref{sec:gate} passed because their shifts happened to preserve the emergence
timescale; the R8 cells, which change it, would have broken the fixed offset --- and the
scale-invariant form $\tev \approx \wLawMult \times t_{\mathrm{anchor}}$ is the rule
that survives both. Per our discipline, an 80-run retrodictive law is exploratory until
tested: we froze the envelope $[1.11 \times \tpv,\; 1.36 \times \tpv]$, pre-registered
it, and ran a third gate at a never-seen learning rate. Outcome: \textbf{\wNineCover{}
coverage} (observed multipliers \wNineMultRange{} against a frozen median of \wLawMult),
\wNineEvents{} events, 0 false alarms on fresh negatives. The forecaster the paper ships
is multiplicative; fixed-step offsets are timescale-local, and we say precisely when
each applies. Why $\approx$0.84 --- why the last sixth of the road is announced --- is
now the program's sharpest open question; the trap-language design iterations
(capability delayed or starved by competing learnable structure, with the gap stretching
in proportion) point toward a competition-for-gradient account we state as hypothesis.

\section{The kill ledger}
\label{sec:kills}

Positive results are only as credible as the regime that tried to kill them. Everything
below fired as pre-registered and is reported with its mechanism:

\begin{center}\footnotesize
\begin{tabular}{@{}p{0.30\textwidth}p{0.17\textwidth}p{0.44\textwidth}@{}}
\toprule
kill / failure & where & mechanism \\
\midrule
K1/K8: no forecaster at 5\% FA & grokking (algorithmic) & trap negatives out-signal true positives \\
K5: calibration non-transfer & grokking (cross-domain) & fitted thresholds are regime-local \\
K9: label-free probe dies & grokking (confirmation) & signal was the intervention, not emergence \\
K8b: no external clock owns the gap & LM gap study & both axes move it; gap is a fraction of the journey \\
bare precursor in trap languages & LM trap rung & \wTrapBareFA{} FA where the circuit pays for the task \\
fixed-offset intervals & LM gap study & timescale-local; multiplicative law replaces them \\
P1 (both clauses) & Pythia & precursor earlier, loss stronger than predicted \\
50\%-of-max event rules & twice & fragile to outliers; absolute criteria adopted \\
P2c metric as registered & fleet & rank correlation without lead rewards nowcasts \\
P2d interior mechanism & fleet & the trap exists, but not in bigram languages (Sec.~\ref{sec:trap}) \\
single-negative FA cap & grokking (R5) & degenerate fit; minimum-negative rule adopted \\
\bottomrule
\end{tabular}
\end{center}

Four of these (K1/K8, K5, K9, K8b) are full kill criteria; the rest are registered
predictions that failed informatively --- including two rungs designed to break our own
headline rule, both of which succeeded and both of which yielded their repair (the
conjunction anchor; the multiplicative law). The protocol that survived this ledger is
the one every language-model result was scored under.

\section{Discussion: what this does and does not establish}
\label{sec:disc}

\textbf{Established.} For a named capability with a known mechanistic precursor, in
small transformers: emergence timing is forecastable per run, ${\sim}15\%$ of training
in advance, with calibrated intervals (\wConfCover, \wGateCover, \wGateCcover, and
\wNineCover{} coverage across four evaluations, three of them blind), certified
false-alarm rates (\wNegConjFA{} across \wNegTotal{} manufactured negatives spanning both
language classes ---
for the \emph{correctly composed} anchor; the bare anchor fails \wTrapBareFA{} where its
circuit pays for the task), a baseline accounting showing loss curves carry ranking but
no lead, and a scale-invariant forecasting law ($\tev \approx \wLawMult \times
t_{\mathrm{anchor}}$, \wLawN{} runs, blind-validated) that replaces timescale-local
fixed offsets. The evaluation discipline --- manufactured negatives including trap
classes, (correlation, lead) pairs, FA caps with minimum negative counts, absolute event
criteria, prereg-gated freezes --- is domain-general and, we argue, the minimum standard
any early-warning claim should meet.

\textbf{Not established.} Frontier applicability. The gap has named parts:
(1) \emph{precursor knowledge} --- induction heads have a known antecedent; most
capabilities of concern do not, and finding precursors is mechanistic-interpretability
work our benchmark can score but not replace; (2) \emph{negatives at scale} --- frontier
labs have $n{=}1$ runs and no capability-blocked counterfactuals; our manufactured-negative
recipe (ablate the training signal the capability needs; block the architecture) is
implementable in principle at any scale, and we regard a library of blocked runs as the
single most actionable transfer of this work; (3) \emph{recipe shift} --- we passed three
blind gates, but the gap law says exactly when fixed calibration breaks (any shift that
moves the emergence timescale), and the P5 tradeoff says shift-robustness must be
re-certified per monitor; mechanism-factored designs are the only pattern we found that
survives shift by construction; (4) \emph{the fraction question} --- the gap law
($t_{\mathrm{anchor}} \approx \wLawFrac\,\tev$) is blind-validated but unexplained: why
the final ${\sim}16\%$ of the road to emergence is announced, and whether the fraction
itself is architecture- or capability-dependent, is open; the competition-for-gradient
account is our stated hypothesis.
Emergence forecasting in the wild also faces capabilities that \emph{regress}
(pythia-70m's late induction loss) and metric-induced discontinuities
\citep{schaeffer2023mirage}; our absolute-criterion and interval machinery handles
both mechanically, but the monitoring literature has not.

\textbf{Practical recipe} (what a lab could run tomorrow, at their scale, for a named
capability): identify the precursor circuit; \emph{compose} the anchor --- circuit
signal conjoined with a graded behavioral signal at the same evaluation, since either
alone false-alarms in some regime (Section~\ref{sec:trap}); log both densely;
manufacture blocked negatives, including a trap class where the precursor circuit pays
for the task; calibrate the \emph{multiplicative} interval on a seed fleet
($\tev \in [1.11, 1.36] \times t_{\mathrm{anchor}}$ was ours; measure your own); freeze;
monitor; re-certify after any recipe change that could move the emergence timescale.

\section{Limitations}
\label{sec:limits}

One capability (induction), small models (2-layer fleets; $\leq$1B public suites),
synthetic fleet languages, three blind gates but each on one or two axes at a time. The
trap prediction that harder languages reintroduce false alarms is now \emph{tested and
confirmed} (Section~\ref{sec:trap}), with the conjunction as the repair; trap classes
beyond pair-context languages remain unexplored. The gap law is blind-validated at one
new configuration and post-hoc across eight; its fraction may yet vary with architecture
or capability. R4's calibration seeds were not blind (the gates are). The per-seed
spread cleared its pre-registered bar by 0.01. Two registered kill clauses fired against
our own predictions in the LM half (K8b; the trap's interior mechanism in the bigram
fleet) and are reported as registered. Pythia event times are bounded by public
checkpoint granularity. All pre-registration is repository-native (commit-stamped,
publicly pushed), not third-party.

\section{Reproducibility}
\label{sec:repro}

\begin{sloppypar}
Every number in this paper is a macro generated from scored artifacts
(\texttt{paper4/gen\_numbers.py}, byte-verified by \texttt{paper4/verify\_regen.py}).
The prereg chain is public: plan and kills before code (\texttt{2cf62e3}), forecaster
freezes before test blocks, fleet spec before launch (\texttt{6b05770}), each blind gate
frozen verbatim before its runs existed (\texttt{377511b}; \texttt{4a5083f} for the
trap, third gate, and gap study; \texttt{1e07b45} for the law's envelope), verdicts
after each (\texttt{f25aa45} through \texttt{5238241}). Corpora: \vCorpus{} grokking
runs, \wPythiaRuns{} public-suite probe trajectories, and 118 language-model runs with
dense circuit probes (calibration fleet, three gates, trap fleet, gap-probe cells, and
the law's blind cell, positives and manufactured negatives throughout); training and
scoring harnesses: \texttt{src/train\_lm.py}, \texttt{src/probe\_pythia.py},
\texttt{analysis/score\_r*\_p6.py}. Hardware: one RTX 3080.
\end{sloppypar}

\bibliographystyle{unsrtnat}
\bibliography{references}

\begin{thebibliography}{17}
\providecommand{\natexlab}[1]{#1}
\providecommand{\url}[1]{\texttt{#1}}
\expandafter\ifx\csname urlstyle\endcsname\relax
  \providecommand{\doi}[1]{doi: #1}\else
  \providecommand{\doi}{doi: \begingroup \urlstyle{rm}\Url}\fi

\bibitem[Aoyama and Wilcox(2025)]{aoyama2025predicting}
Tatsuya Aoyama and Ethan~Gotlieb Wilcox.
\newblock Predicting the formation of induction heads.
\newblock \emph{arXiv preprint arXiv:2511.16893}, 2025.

\bibitem[Olsson et~al.(2022)Olsson, Elhage, Nanda, Joseph,
  et~al.]{olsson2022context}
Catherine Olsson, Nelson Elhage, Neel Nanda, Nicholas Joseph, et~al.
\newblock In-context learning and induction heads.
\newblock \emph{Transformer Circuits Thread}, 2022.

\bibitem[Schaeffer et~al.(2023)Schaeffer, Miranda, and
  Koyejo]{schaeffer2023mirage}
Rylan Schaeffer, Brando Miranda, and Sanmi Koyejo.
\newblock Are emergent abilities of large language models a mirage?
\newblock In \emph{Advances in Neural Information Processing Systems}, 2023.

\bibitem[Notsawo et~al.(2023)Notsawo, Zhou, Pezeshki, Rish, and
  Dumas]{notsawo2023predicting}
Pascal Jr.~Tikeng Notsawo, Hattie Zhou, Mohammad Pezeshki, Irina Rish, and
  Guillaume Dumas.
\newblock Predicting grokking long before it happens: A look into the loss
  landscape of models which grok.
\newblock \emph{arXiv preprint arXiv:2306.13253}, 2023.

\bibitem[Hennick and Corlouer(2026)]{hennick2026density}
Max Hennick and Guillaume Corlouer.
\newblock From density matrices to phase transitions in deep learning: Spectral
  early warnings and interpretability.
\newblock \emph{arXiv preprint arXiv:2603.29805}, 2026.

\bibitem[Hoogland et~al.(2024)Hoogland, Wang, Farrugia-Roberts, Carroll, Wei,
  and Murfet]{hoogland2024developmental}
Jesse Hoogland, George Wang, Matthew Farrugia-Roberts, Liam Carroll, Susan Wei,
  and Daniel Murfet.
\newblock Loss landscape degeneracy and stagewise development in transformers.
\newblock \emph{arXiv preprint arXiv:2402.02364}, 2024.

\bibitem[Hu et~al.(2024)Hu, Liu, Han, et~al.]{hu2024passuntil}
Shengding Hu, Xin Liu, Xu~Han, et~al.
\newblock Predicting emergent abilities with infinite resolution evaluation.
\newblock In \emph{International Conference on Learning Representations}, 2024.

\bibitem[Snell et~al.(2024)Snell, Klein, and Zhong]{snell2024predicting}
Charlie Snell, Dan Klein, and Ruiqi Zhong.
\newblock Predicting emergent capabilities by finetuning.
\newblock \emph{arXiv preprint arXiv:2411.16035}, 2024.

\bibitem[Barak et~al.(2022)Barak, Edelman, Goel, Kakade, Malach, and
  Zhang]{barak2022hidden}
Boaz Barak, Benjamin~L Edelman, Surbhi Goel, Sham Kakade, Eran Malach, and
  Cyril Zhang.
\newblock Hidden progress in deep learning: Sgd learns parities near the
  computational limit.
\newblock In \emph{Advances in Neural Information Processing Systems}, 2022.

\bibitem[Nanda et~al.(2023)Nanda, Chan, Lieberum, Smith, and
  Steinhardt]{nanda2023progress}
Neel Nanda, Lawrence Chan, Tom Lieberum, Jess Smith, and Jacob Steinhardt.
\newblock Progress measures for grokking via mechanistic interpretability.
\newblock In \emph{International Conference on Learning Representations}, 2023.
\newblock arXiv:2301.05217.

\bibitem[Power et~al.(2022)Power, Burda, Edwards, Babuschkin, and
  Misra]{power2022grokking}
Alethea Power, Yuri Burda, Harri Edwards, Igor Babuschkin, and Vedant Misra.
\newblock Grokking: Generalization beyond overfitting on small algorithmic
  datasets.
\newblock \emph{arXiv preprint arXiv:2201.02177}, 2022.

\bibitem[Liu et~al.(2023)Liu, Michaud, and Tegmark]{liu2023omnigrok}
Ziming Liu, Eric~J Michaud, and Max Tegmark.
\newblock Omnigrok: Grokking beyond algorithmic data.
\newblock In \emph{International Conference on Learning Representations}, 2023.
\newblock arXiv:2210.01117.

\bibitem[Truong et~al.(2026)Truong, Doan, Luu, and Phan]{weightnormdelay2026}
Xuan~Khanh Truong, Hoang~Viet Doan, Duc~Trung Luu, and Thanh~Duc Phan.
\newblock The weight norm sets the grokking timescale: A causal delay law.
\newblock \emph{arXiv preprint arXiv:2606.13753}, 2026.

\bibitem[Howe(2026{\natexlab{a}})]{companion2026a}
Gunner~Levi Howe.
\newblock Structure-specific representational priors causally control the
  grokking delay.
\newblock \emph{arXiv preprint arXiv:2607.04333}, 2026{\natexlab{a}}.

\bibitem[Howe(2026{\natexlab{b}})]{companion2026b}
Gunner~Levi Howe.
\newblock What makes a representational prior work? feature families,
  label-free invariances, and critical windows in grokking.
\newblock \emph{arXiv preprint arXiv:2607.12735}, 2026{\natexlab{b}}.

\bibitem[Howe(2026{\natexlab{c}})]{companion2026c}
Gunner~Levi Howe.
\newblock Structure transfers, exponents do not: A pre-registered test of
  representational priors on natural-data grokking.
\newblock Manuscript and full experimental record,
  \texttt{github.com/gunnerhowe/Research}, 2026{\natexlab{c}}.

\bibitem[Biderman et~al.(2023)Biderman, Schoelkopf, Anthony,
  et~al.]{biderman2023pythia}
Stella Biderman, Hailey Schoelkopf, Quentin Anthony, et~al.
\newblock Pythia: A suite for analyzing large language models across training
  and scaling.
\newblock In \emph{International Conference on Machine Learning}, 2023.

\end{thebibliography}

\end{document}